\documentclass{article}
\usepackage{spconf,amsmath,graphicx}
\usepackage{multirow}
\usepackage{graphicx}
\usepackage{subcaption}
\usepackage{comment}
\usepackage{amsfonts}
\usepackage{booktabs}

\title{Two-Stream temporal transformer for video action classification}
\name{Nattapong Kurpukdee and Adrian G. Bors} 
\address{Department of Computer Science,\\ 
University of York, York YO10 5GH, UK\\
{\tt\small nattapong.kurpukdee@york.ac.uk, adrian.bors@york.ac.uk}}
\begin{document}
%
\maketitle
\begin{abstract}
Motion representation plays an important role in video understanding and has many applications including action recognition, robot and autonomous guidance or others. Lately, transformer networks, through their self-attention mechanism capabilities, have proved their efficiency in many applications. In this study, we introduce a new two-stream transformer video classifier, which extracts spatio-temporal information from content and optical flow representing movement information. The proposed model identifies self-attention features across the joint optical flow and temporal frame domain and represents their relationships within the transformer encoder mechanism. The experimental results show that our proposed methodology provides excellent classification results on three well-known video datasets of human activities.
\end{abstract}
\begin{keywords}
Video Transformer, Optical Flow, Two-Stream video processing, Video Action Classification.
\end{keywords}
\section{Introduction}
\label{sec:intro}

Video action classification is an active research area with many applications related to improving the society well-being and security. So far video processing and understanding relied on applying Convolutional Neural Networks (CNNs) to various tasks. Good video classification results have been obtained by multi-stream video processing networks such as the Slow-Fast model \cite{slow-fast} where RestNet \cite{he2016deep} was used as the backbone for extracting multi-stream data, as well as the Busy-Quiet model \cite{BQN}.
However, recently transformers have emerged as a powerful deep learning network relying to their self-attention mechanism. Current state-of-the-art transformers \cite{mvit, mvitv2, Yan_2022_CVPR} achieve better results than CNNs on standard video benchmark databases such as UCF101 \cite{UCF101} and HMDB51 \cite{HMDB51}, Kinetic-400 dataset \cite{kay2017kinetics}, and Something-Something V2 \cite{something}. Multi-view transformers was lately have shown their efficiency in \cite{Yan_2022_CVPR}.

Video action recognition, when compared to image classification, raises many additional challenges posed by the uncertainty with respect to the video capturing conditions, because of the variability of the environment or of the capturing parameters. Moreover, video classification also requires additional training resources including training examples and computational resources. The traditional two-stream architectures \cite{carreira2018quo, feichtenhofer2016convolutional, BQN,8560165, sun2018optical, yue2015beyond},  consider training separate network branches decomposing the video into two different streams. 
Lately, neural networks are increasingly being used for optical flow estimation. The performance of the neural optical flow model is excellent when compared with the output of hand-crafted methods as shown in several studies \cite{dosovitskiy2015flownet, teed2020raft}. 

This research study develops a model for efficiently extracting and using representation features for video classification. Transformers are shown to be able to intrinsically detect the similarities in data. A transformer-based architecture fusing scene and movement features from video through a self-attention mechanism by a video transformer, is proposed in this research study for video classification.
The transformer self-attention mechanism is shown to jointly model efficiently video features from the content defined by sets of frames and the movement from the corresponding optical flow. 

The rest of the paper is organized as follows. In Section~\ref{sec:prior} we provide an overview of the related work on video action recognition using deep network architectures. In Section~\ref{sec:method} we introduce the proposed two-stream video transformer processing approach. The video action classification performance is evaluated in Section~\ref{sec:experimantal_result} followed by the discussion of the results and comparisons with other models. The conclusions are drawn in Section~\ref{sec:conclusion}.

\section{RELATED WORK}
\label{sec:prior}

Traditional video classification research has been successful at developing video descriptors encoding both appearance and motion information \cite{carreira2018quo, feichtenhofer2016convolutional, 8560165, simonyan2014two, sun2018optical,tang2019hallucinating, wang2016temporal, yue2015beyond}. Prior approaches \cite{simonyan2014two} employ two-stream convolution neural networks for action recognition in videos. They show that their network combines single spatial features and temporal optical flow features then using class score fusion for video recognition. Non-local video operators have been studied in \cite{Non-local}. In \cite{feichtenhofer2016convolutional} presents the study of fusion features between the single spatial and temporal optical flow feature.Moreover, in \cite{sun2018optical} the effectiveness by using the optical flow for motion representation was shown. With the optical flow guidance, they can achieve the video action classification. Traditional optical flow such as \cite{horn1981determining} is used to extract motion features and CNNs neural network is a neural network for learning both spatial and temporal features. MoNet \cite{tang2019hallucinating} also provides the optical flow prediction without the heavy optical flow computation and improves the video classification performance. The authors of \cite{nebisoy2021video} also applied optical flow features together with the spatial features and the Long Short-Term Memory (LSTM) CNN \cite{zhu2020comprehensive} were used to achieve video action recognition. In addition, the Slow-Fast model \cite{slow-fast} approach employs local appearance and movement information using various features, which are then fused to produce a global video-level descriptor.

\begin{figure}[t!]
\begin{minipage}[b]{1.0\linewidth}
  \centering
  \centerline{\includegraphics[width=8.5cm]{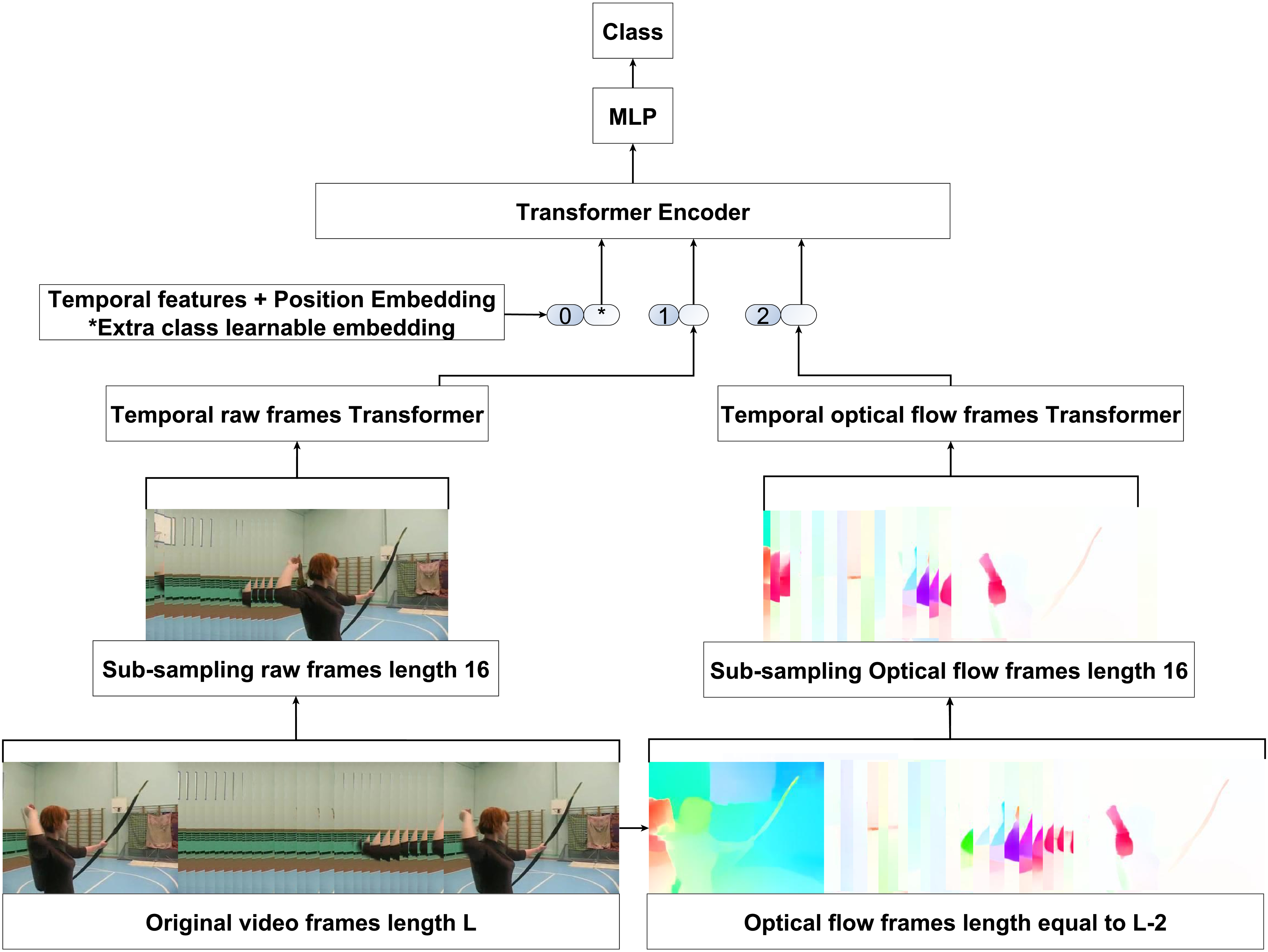}}
\end{minipage}
\caption{The two-stream temporal transformer. The raw frames and optical flow are fed as inputs into the transformer neural network. The extracted features are then processed by the transformer encoder. Finally, a Multi-Layer Perceptron (MLP) is used to classify the video classes.}
\label{fig:overview}
\end{figure}

Currently, the transformer model \cite{attention_is_all_you_need} is increasingly employed in various  video classification tasks \cite{ mvit, mvitv2, VideoSwinTransformer,Yan_2022_CVPR}. The recent successes of the transformer approach for video classification in the multi-scale vision transformer version 1 \cite{mvit} and the corresponding improved processing time version 2 \cite{mvitv2} by using a transformer for encoding video patches into a deep scale data representation have opened an entire new area of applications for deep networks. The multi-view transformer \cite{Yan_2022_CVPR}  uses different frame frequencies for different input views. In addition, \cite{two-stream-transformer-for-long-video} also uses a two-stream spatio-temporal convolutional network employing a spatial-based transformer as well as a temporal transformer which feeds into a separate MLP. Then, both MLPs are merged for the final video classification output.
However, no attempts have been made through using a transformer for identifying self-attention similarities in the content and movement and for optimally fusing motion information and temporal features for characterizing video information. Moreover, a neural network model such as the FlowFormer \cite{huang2022flowformer} and RAFT \cite{teed2020raft}, have shown significant improvements in optical flow estimation over classical hand-crafted movement estimation algorithms.

\section{Methodology}
\label{sec:method}

In this study, we aim to classify videos by using two data streams representing both static scene and movement representations as inputs. The scene content is indicated by a series of image frames, obtained by temporally sub-sampling the videos, thus reducing the information being processed, while the movement is represented by the corresponding optical flow.
We propose to use an efficient transformer backbone which enables the extraction of video features from selected frames and their corresponding optical flows. The transformer encoder finds self-similarities in such features spanning both fused content and movement latent spaces and fuses such features eventually enabling data characteristics recognition.  We present the overview of our proposed model in Fig~\ref{fig:overview}.

\subsection{Optical flow as movement representation}

First, we consider a neural network for predicting the optical flow instead of using a classical handcraft extraction. In general, the flow will indicate the displacement of every single pixel, or of groups of pixels, between a pair of images. Given a pair of consecutive RGB images, $I_1$, $I_2$, the model will estimate a dense displacement field ($f^1$,$f^2$) which maps each pixel ($u$, $v$) in $I_1$ to its corresponding coordinates in $I_2$ as in the following~:
\begin{equation}
(u^\prime, v^\prime) = (u + f^1(u), v + f^2(v))
\label{eq:flow1}
\end{equation}

\begin{figure}[ht!] 
  \begin{subfigure}[b]{0.5\linewidth}
    \centering
    \includegraphics[height=3cm,width=0.9\linewidth]{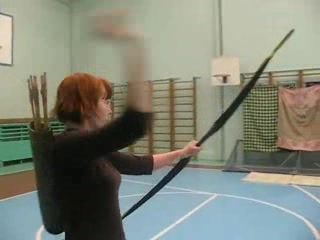} 
    \caption{UCF101 Input image} 
    \label{fig:flowa} 
  \end{subfigure}
  \begin{subfigure}[b]{0.5\linewidth}
    \centering
    \includegraphics[height=3cm,width=0.9\linewidth]{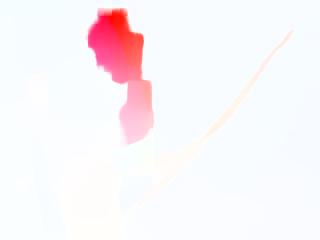} 
    \caption{UCF101 Optical flow image} 
    \label{fig:flowb} 
  \end{subfigure} 
  \begin{subfigure}[b]{0.5\linewidth}
    \centering
    \includegraphics[height=3cm,width=0.9\linewidth]{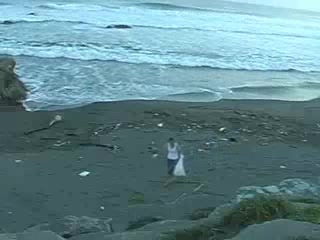} 
    \caption{HMDB51 Input image} 
    \label{fig:flowc} 
  \end{subfigure}
  \begin{subfigure}[b]{0.5\linewidth}
    \centering
    \includegraphics[height=3cm,width=0.9\linewidth]{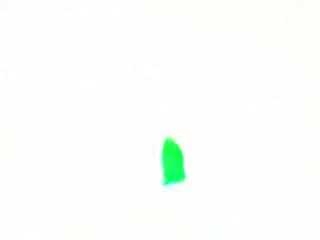} 
    \caption{HMDB51 Optical flow image} 
    \label{fig:flowd} 
  \end{subfigure} 
  \begin{subfigure}[b]{0.5\linewidth}
    \centering
    \includegraphics[height=3cm,width=0.9\linewidth]{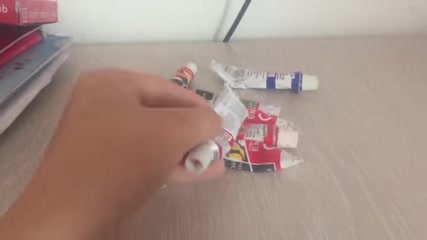} 
    \caption{SSv2 Input image} 
    \label{fig:flowe} 
  \end{subfigure}
  \begin{subfigure}[b]{0.5\linewidth}
    \centering
    \includegraphics[height=3cm,width=0.9\linewidth]{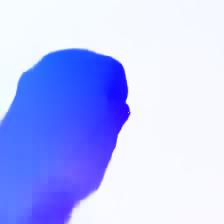} 
    \caption{SSv2 Optical flow image} 
    \label{fig:flowf} 
  \end{subfigure} 
  \caption{Examples of frames and their corresponding optical flow prediction, where the optical flow is estimated by the RAFT model \cite{teed2020raft} on videos from UCF101, HMDB51, and SSv2 dataset. Each direction in the flow is mapped to a RGB color, according to the vector orientation.}
  \label{fig:opticalflow} 
\end{figure}

In the first stage, of preparing the training data, we sample a sequence of image frames from each given video sequence. Two consecutive frames will be used as a  pair for predicting the flow. We consider RGB images for visualizing the optical flow, where the colour indicates the direction of the local movement vector. Examples of various optical flow representations are illustrated in Fig~\ref{fig:flowb},~\ref{fig:flowd},~\ref{fig:flowf} for the corresponding frames from Fig~\ref{fig:flowa},~\ref{fig:flowc},~\ref{fig:flowe}, respectively. This demonstrated the clear movement of objects. Consequently, we can apply the process of the movement data in the same way as the sampled video frames.

\subsection{Video Transformer}

In this section, we describe the video transformer operation, which is used to model the self-attention mechanisms within an increasing spatio-temporal resolution representation of video frames and corresponding movement flow. Both RGB and optical flow inputs are of size $T \times H \times W \times C$, where $T$ corresponds to the number of frames $H$ corresponds to the height $W$ corresponds to the width and $C$ corresponds to the number of channels. After ﬂattening each image and corresponding optical flow. Segmented spatio-temporal tensor defined patches ${\bf X} \in \mathbb{R}^{T \times H \times W \times C }$ are then processed by the transformer. Following the attention mechanism, the transformer projects the input ${\bf X}$ into the query tensor ${\bf Q} \in \mathbb{R}^{T \times H \times W \times C }$, key tensor ${\bf K} \in \mathbb{R}^{T \times H \times W \times C }$, and value tensor 
${\bf V} \in \mathbb{R}^{T \times H \times W \times C }$. The input of the transformer consists of query and keys of dimension $d_k$, and values of dimension $d_v$. Then compute the dot products of the query with all keys, divide each by $\sqrt{d_k}$, and apply a softmax function to obtain the weights on the values as described in~:
\begin{equation}
Attn({\bf Q},{\bf K},{\bf V}) = softmax \left( 
\frac{{\bf Q}{\bf K}^T}{\sqrt{d_k}}\right) {\bf V} .
\label{eq:attention}
\end{equation}

The self-attention mechanism of the transformer is deployed in the spatio-temporal representation of the video space, using pooling operations in order to reduce the latent tensor dimension instead of considering reducing the input pixel tensor dimension. Reducing the latent tensor dimension speeds up the video processing. Finally, the transformer encodes each input into a certain feature vector. Various types of transformers are considered in the experimental results including the Multi-scale vision transformer (MViTv2) \cite{mvitv2} and the video Swin transformer \cite{VideoSwinTransformer}, each achieving remarkable successes according to their original research studies. 

\subsection{Two-Stream Fusion and Classification}

Unlike in other applications, such as when using CNN networks, the transformer encoder enables the self-attention mechanism on fused features instead of simply concatenating them into a single vector. 
The two-stream architecture encodes RGB frame and optical flow frame into output features vectors ${\bf O}_r$ and ${\bf O}_f$, respectively. 
Each output vector considers a $D_o$ dimensional output tensor ${\bf O}$ of length $L_o$, ${\bf O}_r \in \mathbb{R}^{L_o \times D_o}$, ${\bf O}_f \in \mathbb{R}^{L_o \times D_o}$. The output vector with the extra class learnable vector ${\bf X}_{class} \in \mathbb{R}^{L_o \times D_o}$ is mapped with the positional embedding ${\bf E}_{pos}$ for the transformer input like in the ViT \cite{dosovitskiy2020image} and described in the following~:
\begin{equation}
{\bf X} = [ {\bf X}_{class}; {\bf O}_r^1{\bf E}; {\bf O}_f^2 {\bf E}; ] + {\bf E}_{pos}.
\label{eq:fuse_transformer_input}
\end{equation}
 
Where ${\bf E} \in \mathbb{R}^{L_o \times D_o}$, ${\bf E}_{pos} \in \mathbb{R}^{(N+1) \times D_o}$, and $N$ is the number of inputs equal to 2. 
Considers a $D$ dimensional input tensor ${\bf X}$ of sequence length $L$, ${\bf X} \in \mathbb{R}^{L \times D}$.
Instead of performing a single attention function with $D$-dimensional keys, values, and queries, we apply the Multi-Head Attention (MHA) in order to linearly project the queries, keys, and values $h$ times with different, learned linear projections to $d_k$, $d_k$ and $d_v$ dimensions, respectively. 
Following the MHA, we project the input ${\bf X}$ into the query tensor ${\bf Q} \in \mathbb{R}^{L \times D}$, key tensor ${\bf K} \in \mathbb{R}^{L \times D}$, and value tensor ${\bf V} \in \mathbb{R}^{L \times D}$. 
On each of the projected versions of query, keys, and values we then perform the attention function in parallel, yielding $d_v$ output values. These are concatenated and once again projected, resulting in the final values given by~:
\begin{equation}
\begin{split}
MHA({\bf Q},{\bf K},{\bf V}) = Concat(head_1,...,head_h){\bf W}^o \\
head_i = Attn({\bf Q}{\bf W}_i^Q, {\bf K}{\bf W}_i^K,
{\bf V}{\bf W}_i^V)
\end{split}
\label{eq:mha}
\end{equation}
where ${\bf W}_i^Q \in \mathbb{R}^{L \times D}$, ${\bf W}_i^K \in \mathbb{R}^{L \times D}$, ${\bf W}_i^V \in \mathbb{R}^{L \times D}$, and 
${\bf W}^{o} \in \mathbb{R}^{hd_v \times D}$. 
In this work, we employ $h$ = 8 parallel attention layers, or heads. For each of these, we use $d_k$ = $d_v$ = $D/h$ = 96.
For the classification module, we apply the MLP network consisting of three layers starting with a normalization layer followed by a dropout layer. Then the final layer is a fully-connected linear layer for the classifier.

\section{Experimental results}
\label{sec:experimantal_result}

In this section, we first discuss the details of the datasets UCF101, HMDB51, and Something-Something V2 used in the experiments. Then we discuss the implementation details followed by the experimental results and their discussion.

\subsection{Datasets and Tasks}

We evaluate our proposed framework on UCF101 \cite{UCF101} and HMDB51 \cite{HMDB51} which are standard datasets for action recognition tasks. The UCF101 dataset consists of 13,320 videos from 101 classes and provides three splits of training, validation, and testing datasets. We randomly select 10\% of each training set as the validation set. The HMDB51 dataset consists of 6,766 video examples from 51 action classes and also provides three splits for training and testing datasets. We average the classification accuracy over all three splits to evaluate the video classification results on either UCF101 or HMDB51 dataset. Meanwhile, Something-Something V2 dataset is a large-scale dataset, which contains 220,847 video examples in total from 174 action classes. We consider 168,913 videos for training and 27,157 for testing, while 24,777 videos represent the validation set. Something-Something V2 is characterized by a diversity of movement information, including movements that are rather specific to specialized activities. The information about the three databases used in the experiments is provided in Table~\ref{tab:data-statistics}. 

\begin{table}[ht!]
\resizebox{\linewidth}{!}{%
\begin{tabular}{lccc}
\hline
\multirow{2}{*}{Datasets}  & \multicolumn{3}{c}{Video Per Task}   \\ \cline{2-4}
                                                                & \multicolumn{1}{c}{Train} & \multicolumn{1}{c}{Val} & \multicolumn{1}{c}{Test}                                    \\ \hline
HMDB51 Fold-1,2,3        & 3,570    & 1,666   &  1,530        \\
 \hline
UCF101 Fold-1             & 7,629      & 1,908    & 3,783                                                         \\
UCF101 Fold-2                                 & 7,668      & 1,918    & 3,734                                                                             \\
UCF101 Fold-3                                 & 7,699                          & 1,925                        & 3,696                                                                                                         \\ \hline
Something-Something V2              & 168,913      & 24,777    & 27,157                                                                                                                                 \\ \toprule
\end{tabular}%
}
\caption{Data Statistics. We consider 7 splits for video classification tests.}
\label{tab:data-statistics}
\end{table}

\subsection{Implementation Details}

We consider the RAFT optical flow model \cite{teed2020raft} for predicting the movement between two images from a sequence of video frames. We treat the data of the optical flow as the same as the RGB frame. We map the optical flow into RGB representation by using the flow\_to\_image function provided by the torchvision utility. Moreover, we consider the Multiscale Vision
Transformers for Classification and Detection (MViTv1) \cite{mvit}, Improved Multiscale Vision Transformers for Classification and Detection (MViTv2) \cite{mvitv2} or the Video Swin Transformer (Swin) \cite{VideoSwinTransformer}, as the backbone for temporal feature extraction. These video transformers are pre-trained on Kinetic-400 dataset \cite{kay2017kinetics}. For all transformer architectures, for the sake of using lower computational resources, we decided to use the smallest model architecture called MViTV1-B, MViTV2-S, and Swin-S. Because these transformer models are designed for a small dataset their operation processing is rather quick. These three models use as input videos of 16 frames of size $224 \times 224 \times 3$. The video input and optical flow input are compressed at the final layer of the channel encoder to 768 channel features.

For the data pre-processing, we follow the protocol for each transformer. For the MViT transformer, we consider a random crop video of $ 224 \times 224$ pixels. Then the pixel values are re-scaled to [0, 1]. After that, the values are normalized using the mean=[0.45, 0.45, 0.45] and standard deviation as std=[0.225, 0.225, 0.225]. For the video Swin transformer, we follow the pre-processing by normalized using mean=[0.485, 0.456, 0.406] and std=[0.229, 0.224, 0.225]. We implement our models using PyTorch with a single GPU Tesla V100 32G on a POWER9 architecture. For each dataset, we train a neural network for a maximum of 200 epochs with a small batch size of 8. We optimize the model using Adam \cite{kingma2014adam} with a learning rate of 0.0002 and dropout at 0.5. A single global transformer encoder will set the number of heads at 8 while the dimension of the feedforward network model is set at 1024. For training, we consider the Cross-entropy loss function. In addition, to avoid over-fitting the model, we apply the early stopping strategy to monitor the validation loss rate. If the validation loss rate does not improve for 10 epochs we would stop the training and start the evaluation. 

\subsection{Video classification results}

In this section, we present the experimental results of our proposed video action classification.  For the evaluation results, we consider the Top-1 Accuracy (Top-1 ACC) representing the classification accuracy of the model. We present the results on UCF101, HMDB51, and Something-Something V2 which are the standard action recognition datasets.

\textbf{Baselines} We retrain and evaluate three widely used state-of-the-art methods as baselines including MViTv1 \cite{mvit}, MViTv2 \cite{mvitv2} and Swin-S \cite{VideoSwinTransformer} in the same environment without any data augmentation. We compare these three baseline training scenarios with our proposed method. Moreover, we also compare with other two-stream based video classification models, which splits the given data in the same manner.

\subsubsection{Video classification performance on UCF101 and HMDB51} 

First, we report the results on the UCF101 and HMDB51 datasets as these two databases are extensively used in video action classification benchmarks. The average results across three splits when considering training on UCF101 and HMDB51 are provided in Table~\ref{tab:ucf101_and_hmdb51_result}, where we compare with baselines and state-of-the-art video classification models. 
In Table~\ref{tab:ucf101_and_hmdb51_result} we group the methods used in the tests according to the database used for pre-training the model. We found that our proposed method shows a significant improvement when compared to the baselines and equivalent to the state-of-the-art models on the UCF101 dataset. Moreover, the results on the HMDB51 dataset show that our model has the best accuracy when compared with other methods. Our model can achieve up to 10.9\% increase when considering UCF101, and up to 25.92\% increase when considering HMDB51, compared to the nearest baseline.

\begin{table}[ht!]
\resizebox{\linewidth}{!}{%
\begin{tabular}{lccc}
\hline
\multirow{2}{*}{Classifier} &
\multirow{2}{*}{Pre-trained} & 
\multicolumn{2}{c}{Average Top-1 Acc} \\ \cline{3-4} 
                            &  & \multicolumn{1}{c}{UCF101} & \multicolumn{1}{c}{HMDB51}       \\ \hline
Two-Stream CNNs  \cite{simonyan2014two}             &     ImageNet             &   88.00\%           &   59.40\% \\
OFF  \cite{sun2018optical}             &      -            &   96.00\%           &   74.20\% \\
Two-Stream CNNs  \cite{feichtenhofer2016convolutional}             &  ImageNet                &   93.50\%           &   69.20\% \\
Two-Stream I3D \cite{carreira2018quo} &        -          &     93.40\%           &   66.40\%\\
Two-Stream I3D \cite{carreira2018quo} &      Imagenet+Kinetics 400            &     \textbf{98.00}\%           &   80.70\%\\
Two-Stream+LSTM \cite{yue2015beyond} &      -            &    88.60\%           &   -\\
Two-Stream C3D \cite{8560165} &      -            &     91.40\%           &   -\\
Two-Stream TSN \cite{wang2016temporal} &      -            &    94.00\%           &   68.50\% \\
Three-Stream TSN \cite{wang2016temporal} &      -            &     94.20\%           &    69.40\% \\ 
TDD+iDT  \cite{wang2015action}             &     -              &   91.50\%           &   65.90\%            \\ 
LTC+iDT  \cite{varol2017long}             &     -              &   91.70\%           &   64.80\%            \\ 
ST-ResNet + IDT \cite{feichtenhofer2016spatiotemporal}             &     -              &   94.60\%           &   70.30\%            \\ 

\hline
MViTv1-B finetune (our baseline)\cite{mvit}         &  Kinetics-400        &   89.66\%           &    66.75\%      \\ 
MViTv2-S finetune (our baseline) \cite{mvitv2}       & Kinetics-400      &   92.11\%                  &    73.59\%      \\ 
Swin-S finetune (our baseline)\cite{VideoSwinTransformer}    & Kinetics-400       &   82.64\%                  &    57.47\%          \\ \cline{3-4}
Our (MViTv2-S based)         & Kinetics-400      &   \textbf{93.54}\%         &    \textbf{83.39\%}           \\ 
\hline
\end{tabular}%
}
\caption{Video classification results on UCF-101 and HMDB-51, representing the average over 3 splits.
}
\label{tab:ucf101_and_hmdb51_result}
\end{table}

\subsubsection{Video classification performance on SSv2} 

Second, we have assessed the proposed methodology on the Something-Something V2 as this dataset is far more complex than the others used in video action classification. The results are provided in Table~\ref{tab:ssv_result} where we compare with baselines and state-of-the-art models. The methods are grouped according to the database used for pre-training the model. We can observe that the proposed method provides a significant improvement when compared to the baselines and can achieve up to 6.82\% increase in the classification performance when compared to the baseline without any data augmentation.

\begin{table}[ht!]
\resizebox{\linewidth}{!}{%
\begin{tabular}{lcc}
\hline
\multirow{2}{*}{Classifier} &
\multirow{2}{*}{Pre-trained} & 
Top-1 Acc \\ \cline{3-3} 
                            &  & \multicolumn{1}{c}{SSv2}        \\ \hline
TimeSformer-HR \cite{bertasius2021space} & ImageNet-21K & 62.5\% \\
SlowFast R101, 8×8 \cite{feichtenhofer2019slowfast} &  Kinetics-400       & 63.1\%\\
TSM-RGB \cite{lin2019tsm}             &    ImageNet-1K+K400               &  63.3\%\\
MSNet \cite{kwon2020motionsqueeze}    &    ImageNet-1K                    &  64.7\%\\
TEA \cite{li2020tea}                  &    ImageNet-1K                    &  65.1\%\\
ViT-B-TimeSformer \cite{bertasius2021space}   &    ImageNet-21K           &  62.5\%\\
ViT-B               &    ImageNet-21K                   &  63.5\%\\
MViTv1-B, 16×4 \cite{mvit}       &  Kinetics-400  &  64.7\%\\
MViTv2-S, 16×4 \cite{mvitv2}     &  Kinetics-400            &  68.2\%\\
Swin-B \cite{VideoSwinTransformer}         & Kinetics-400             & \textbf{69.6}\%  \\ 
bLResNet, 32x2 \cite{fan2019more}        &     SSv2      & 67.1\%  \\ \hline
MViTv1-B finetune (our baseline)\cite{mvit}                       &  Kinetics-400    &  50.31\% \\ 
MViTv2-S finetune (our baseline) \cite{mvitv2}         &  Kinetics-400           & 53.36\%  \\ 
Swin-S finetune (our baseline)\cite{VideoSwinTransformer}  & Kinetics-400             & 49.56\% \\ \cline{3-3}
Our (MViTv2-S based)             & Kinetics-400             & \textbf{56.38\%}   \\ 
\hline
\end{tabular}%
}
\caption{Video classification performance comparison to baseline and the state-of-the-art methods on Something-Something V2.
}
\label{tab:ssv_result}
\end{table}

\subsection{Discussions}

\subsubsection{Data augmentaion}
To observe that the performance of the proposed method relies on limited resources, such as employing the smallest video processing architecture available without using any data augmentation strategy in both the baseline model and our two-stream model. The results show that our proposed method can achieve better video accuracy when compared with the baseline in all datasets. However, in future studies, our proposed model will require a data augmentation strategy when considering learning complex data as same as described in the state-of-the-art techniques \cite{slow-fast, mvit, mvitv2, VideoSwinTransformer} such as flipping, Mixup \cite{zhang2017mixup} with CutMix \cite{yun2019cutmix}, Random Erasing \cite{cubuk2020randaugment},and Random Augment \cite{zhong2020random}.

\subsubsection{Optical flow}
We produce the optical-flow prediction based on a neural network model instead of hand-crafted prediction. The model trained on FlyingChairs \cite{dosovitskiy2015flownet} + FlyingThings3D \cite{mayer2016large} dataset. Both datasets are synthetic and not realistic, actually far in thir appearance from human activity video. A good optical flow prediction can be shown in Fig~\ref{fig:opticalflow} with a clear moving object direction and position when mapped into RGB color. This is clear that the optical flow model learned from the synthetic data can perform well on realistic activity data. However, we found some weaknesses in the optical flow prediction such as producing noisy flow images in some cases. Thus, a better movement estimation could improve the classification results. Moreover, the computational cost for estimation optical flow by neural network model is rather high and we have to balance the accuracy of the optical flow and the processing computation required.

\section{Conclusion}
\label{sec:conclusion}

In this paper, we propose a two-stream transformer architecture for video classification. A multi-scale video transformer is used to extract self-attention features from combining sampled image frames and corresponding optical flow. For the two-stream fusion, we apply a pre-trained transformer encoder to learn the relation between the original frame and the optical flow frame. By using a powerful transformer architecture the proposed approach can harness the power of defining well the inter-relationships among the spatio-temporal features from the original image frame and optical flow frame. The experimental results indicate significant improvement when compared with the baselines and other two-stream video processing models. We plan to use this architecture for video continual learning which has many potential applications.
 
\bibliographystyle{IEEEbib}
\bibliography{strings}

\end{document}